\documentclass{article}

\usepackage{PRIMEarxiv}

\usepackage{natbib}
\usepackage[utf8]{inputenc} 
\usepackage[T1]{fontenc}    
\usepackage{hyperref}       
\usepackage{url}            
\usepackage{booktabs}       
\usepackage{amsfonts}       
\usepackage{nicefrac}       
\usepackage{microtype}      
\usepackage{lipsum}
\usepackage{fancyhdr}       
\usepackage{graphicx}       
\graphicspath{{media/}}     

\newcommand{\setpdfpagesize}[2]{%
  \clearpage
  \pdfpagewidth=#1\relax
  \pdfpageheight=#2\relax
  \paperwidth=#1\relax
  \paperheight=#2\relax
  \newgeometry{paperwidth=#1, paperheight=#2, margin=.2in}%
  \thispagestyle{empty}
}

\newcommand{\resetpdfpagesize}{%
  \clearpage
  \restoregeometry
  \pdfpagewidth=\paperwidth
  \pdfpageheight=\paperheight
}

\usepackage{zi4}
\usepackage{amsmath}
\usepackage{amssymb}
\usepackage{xcolor}
\usepackage{array}
\usepackage{multirow}
\usepackage{pdflscape}
\usepackage{parskip}

\usepackage{listings}
\newcommand{\codeitem}[1]{\lstinline|#1|}
\newcommand{\codeitemfoot}[1]{{\footnotesize\lstinline|#1|}}

\title{Graphical Design of Interpretable Architectures
}

\author{
  Pietro Barbiero \\
  IBM Research \\
  Zurich\\
  \texttt{pietro.barbiero@ibm.com} \\
}

\begin{document}
\maketitle

\begin{abstract}
Designing, implementing, and comparing interpretable architectures requires a formal language to represent them. The most common representations fall short in one of two ways. Symbolic equations give no global view of an architecture at a glance. Probabilistic graphical models and flowcharts do not describe actual tensor manipulations, thus hiding key insights and limiting reproducibility.
To close this gap, we introduce a graphical notation for designing interpretable AI architectures, adapted from Penrose tensor notation. This graphical notation gives a global view of an architecture and maps one to one onto PyTorch \codeitem{einsum} code. We first use this notation to describe architectures that are interpretable by construction, including concept bottlenecks, sparse probes, prototype networks, neural additive models, and mixtures of linear models. We then diagram the key architectural components of Steerling-8B, a frontier interpretable language model. The diagram yields global insights into the architecture (e.g., showing that Steerling is a residual model), a geometric interpretation of each individual operation, and a direct translation into 33 lines of PyTorch code.
\end{abstract}


\section{Introduction}
Frontier AI models manipulate high-dimensional objects called tensors. For this reason, to understand, design, or implement these models, we must think in high-dimensional terms. As frontier models are composed of many such operations, intuitive and formal representations of their tensor manipulations are key to understanding, comparing, and designing state-of-the-art architectures.

Common formal representations, such as symbolic equations, may limit global insights on the architecture at a glance. As an example, the expression
$$f(x_k,w_{kri},t_{krij}) = \sum_i \sigma(x_{k}w_{kri}) t_{krij}$$
hides that the tensor manipulations act on individual features $k$ of $x$ independently. We must carefully work through the whole expression to see this. This process requires time, effort, and it is intrinsically prone to errors. Graphical notations, such as probabilistic graphical models and flowcharts, are commonly used specifically to compensate the shortcomings of symbolic representations and convey immediate insights:
\begin{figure}[!h]
    \centering
    \includegraphics[width=0.6\linewidth]{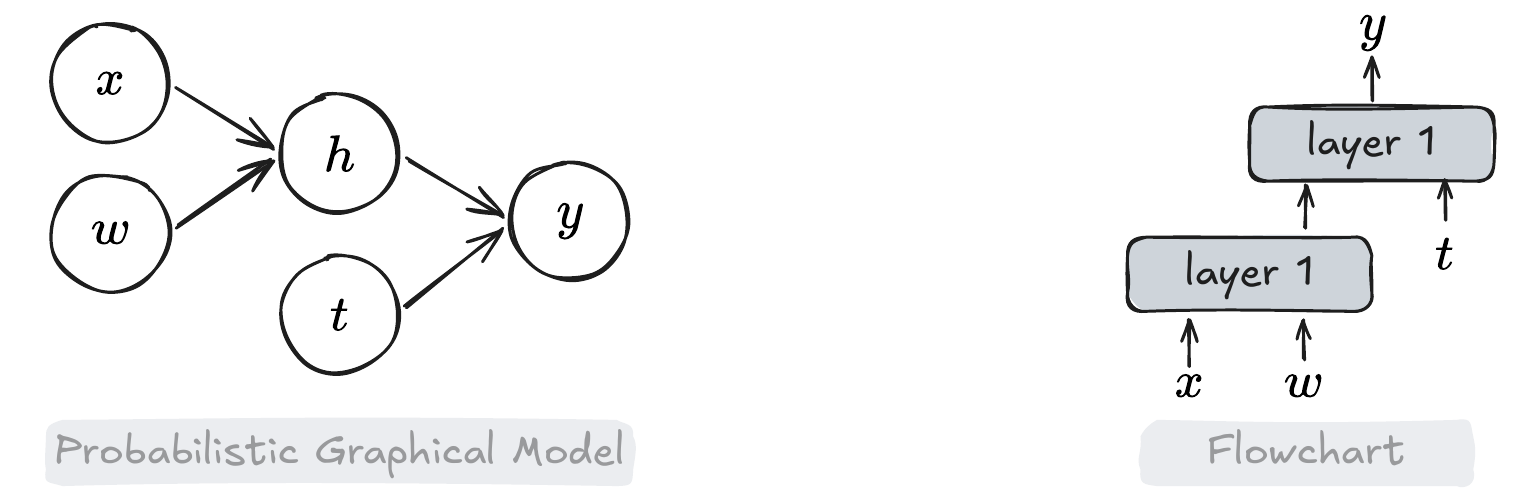}
\end{figure}

However, while these diagrams provide a high-level overview of the architecture, they do not specify how the architecture should be concretely implemented thus obfuscating important innovations and limiting reproducibility.

To solve this, we introduce a graphical notation, adapted from Penrose tensor notation, for designing interpretable AI architectures that enables global insights and maps one to one onto PyTorch \codeitem{einsum} code. While prior work has used a similar notation to analyse language models and mechanistic interpretability, we use this notation to describe architectures that are interpretable by construction, including concept bottlenecks, sparse probes, prototype networks, neural additive models, and mixtures of linear models. As a case study, we diagram the full architecture of Steerling-8B, showing the advantages of such notation in practice.

\section{Graphical Einstein-inspired notation in PyTorch}
A well-known graphical notation for tensor operations, originating in physics, is Penrose or tensor-network notation~\citep{penrose1971applications}. Recent work has adapted it to analyse language models and mechanistic interpretability~\citep{taylor2024introduction,elhage2021mathematical}. To our knowledge, no prior work has used it to analyse models that are interpretable by construction.

We first introduce the fragment of Penrose notation we need. We then use it to analyse the key operations behind frontier interpretable AI models.

\subsection{Tensors}
In this work, a tensor of order $k$ is an array with $k$ indices, $T \in \mathbb{R}^{n_1 \times \dots \times n_k}, n_i,k\in \mathbb{N}$. A scalar has order $0$, a vector order $1$, a matrix order $2$, and so on. The table below lists common tensors, how to generate each at random in PyTorch, and its geometric meaning. In our diagrams, a tensor is a circle with ``legs''. Each leg stands for one geometric dimension, that is, one array index.

\begin{table}[h]
\centering
\resizebox{.65\columnwidth}{!}{%
\renewcommand{\arraystretch}{2.}
\begin{tabular}{
  >{\raggedright\arraybackslash}m{1.5in}
  >{\centering\arraybackslash}m{2.6in}
  >{\centering\arraybackslash}m{.8in}
  >{\centering\arraybackslash}m{.8in}
}
\textbf{Name / op} & \textbf{Diagram} & \textbf{Array shape} & \textbf{Algebraic} \\
\hline
 
Scalar \newline \codeitemfoot{s = torch.randn(1)}
&
\multirow[c]{1}{2.6in}{
    \centering
    \includegraphics[width=1.\linewidth]{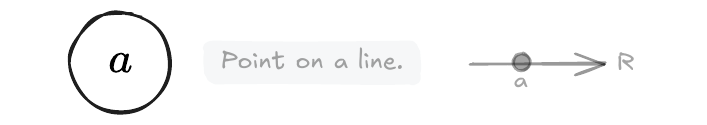}
}
&
{\renewcommand{\arraystretch}{1}$\left[
\begin{smallmatrix} 
\cdot
\end{smallmatrix}\right]$}
&
$s \in \mathbb{R}$
\\ [.3in]
\hline
 
Vector \newline \codeitemfoot{v = torch.randn(n)}
&
\multirow[c]{1}{2.6in}{
    \centering
    \includegraphics[width=1.\linewidth]{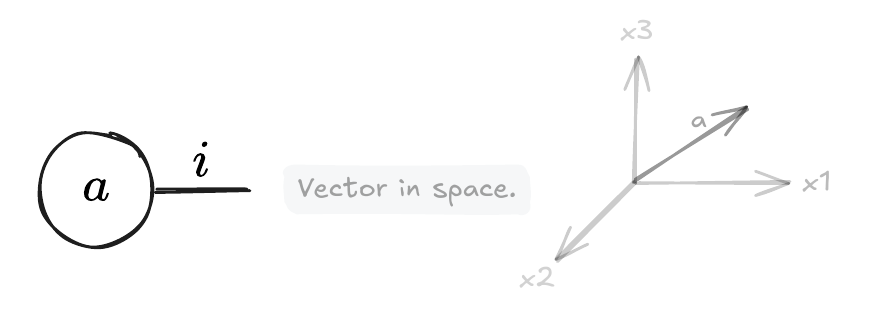}
}
&
{\renewcommand{\arraystretch}{1}$\left[\begin{smallmatrix}\cdot\\\cdot\\\cdot\end{smallmatrix}\right]$}
&
$v \in \mathbb{R}^n$
\\ [.8in]
\hline
 
Matrix \newline \codeitemfoot{A = torch.randn(m, n)}
&
\multirow[c]{1}{2.6in}{
    \centering
    \includegraphics[width=1.\linewidth]{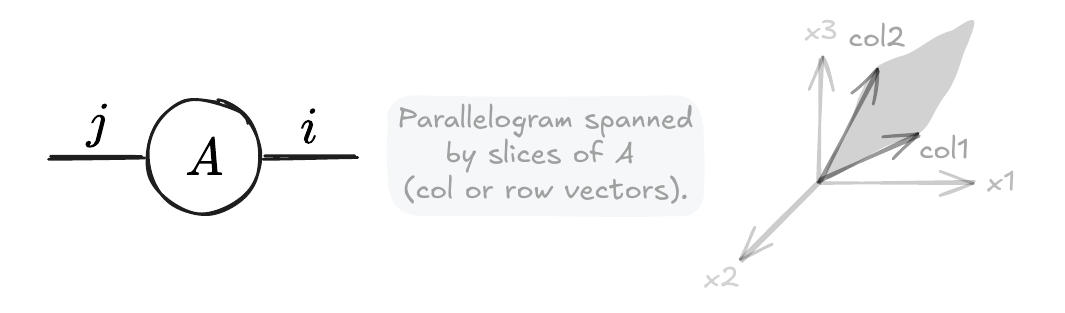}
}
&
{\renewcommand{\arraystretch}{1}$\left[
\begin{smallmatrix} 
\cdot&\cdot\\ 
\cdot&\cdot\\ 
\cdot&\cdot 
\end{smallmatrix}\right]$}
&
$A \in \mathbb{R}^{m \times n}$
\\ [.7in]
\hline
 
3-Tensor \newline \codeitemfoot{T = torch.randn(a, b, c)}
&
\multirow[c]{1}{2.1in}{
    \centering
    \includegraphics[width=1.\linewidth]{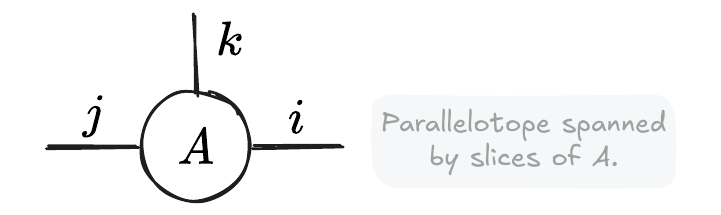}
}
&
--
&
$T \in \mathbb{R}^{a \times b \times c}$
\\ [.5in]
\hline
\end{tabular}
}
\end{table}

\subsection{Operations}
AI models manipulate tensors using tensor operations. Tensor operations admit a formal notation known as \emph{Einstein-inspired notation for operations}~\citep{einstein1916grundlage} or ``Einops''~\citep{rogozhnikov2022einops}. PyTorch supports two main Einops: \codeitem{rearrange}, which reorders tensor's axes, and \codeitem{einsum}, which combines multiple tensors. Given their expressivity, mastering Einops, and \codeitem{einsum} above all, is one of the most useful skills for designing and implementing AI architectures.

Unfortunately, PyTorch \codeitem{einsum} is hard to parse at a glance. It is not the best tool for designing, comparing, or reasoning about tensor operations. It does, however, map one to one onto Penrose graphical notation. This visual notation makes complex tensor operations clear, formal, and unambiguous.

The convention works as follows~\citep{einstein1916grundlage,penrose1971applications}. To combine two tensors, we connect legs that share a label; a shared label marks the same geometric dimension/index. A connected pair of legs is contracted: we multiply the two tensors' entries together for each value of that shared index, then sum over the index. This leaves ``cancels'' the leg out. Any legs left unconnected are ``free'' legs, and they become the indices of the result.

We can turn each diagram directly into PyTorch code with \codeitem{einsum}, using the convention \codeitem{einsum('legs\_of\_input\_tensor\_A,legs\_of\_input\_tensor\_B->legs\_of\_output\_tensor', A, B)}. This lets us design a tensor operation as a diagram, get the diagram's clarity, and then convert it straight into working PyTorch code.

The table below lists the most common tensor operations. These form the building blocks of the tensor manipulations used in frontier interpretable models.

\begin{table}[!h]
\centering
\resizebox{\columnwidth}{!}{%
\renewcommand{\arraystretch}{2.}
\begin{tabular}{
  >{\raggedright\arraybackslash}m{1.8in}
  >{\centering\arraybackslash}m{2.8in}
  >{\centering\arraybackslash}m{1.3in}
  >{\centering\arraybackslash}m{1.2in}
}
\textbf{Name / op} & \textbf{Diagram} & \textbf{Array shape} & \textbf{Algebraic} \\
\hline
 
Scalar multiplication \newline \codeitemfoot{h = torch.einsum(',->', v, w)}
&
\multirow[c]{1}{2.8in}{
    \centering
    \includegraphics[width=1.\linewidth]{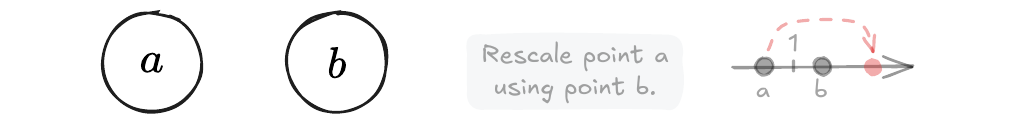}
}
&
$\cdot \times \cdot = \cdot$
&
$h = a b$
\\ [.3in]
\hline

Element-wise product \newline \codeitemfoot{h = einsum('i,i->i', a, b)}
&
\multirow[c]{1}{2.8in}{
    \centering
    \includegraphics[width=1.\linewidth]{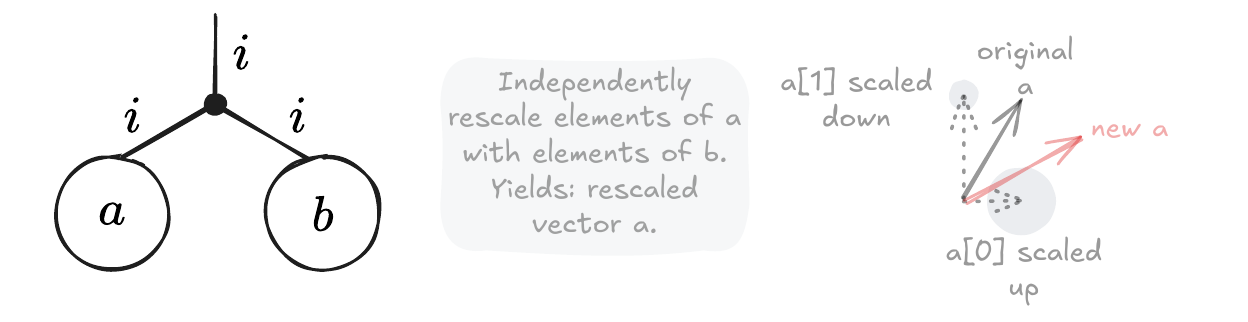}
}
&
{\renewcommand{\arraystretch}{1}$\left[\begin{smallmatrix}\cdot\\\cdot\\\cdot\end{smallmatrix}\right] \odot \left[\begin{smallmatrix}\cdot\\\cdot\\\cdot\end{smallmatrix}\right] = \left[\begin{smallmatrix}\cdot\\\cdot\\\cdot\end{smallmatrix}\right]$}
&
$h_i = a_i b_i$
\\ [.7in]
\hline

Dot product \newline \codeitemfoot{h = einsum('i,i->', a, b)}
&
\multirow[c]{1}{2.8in}{
    \centering
    \includegraphics[width=1.\linewidth]{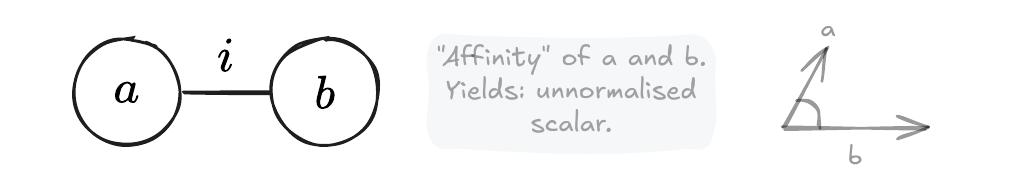}
}
&
{\renewcommand{\arraystretch}{1}$\left[\begin{smallmatrix}\cdot&\cdot&\cdot\end{smallmatrix}\right]\left[\begin{smallmatrix}\cdot\\\cdot\\\cdot\end{smallmatrix}\right] = \cdot$}
&
$h = \sum_i a_i b_i$
\\ [.4in]
\hline
 
Squared norm \newline \codeitemfoot{h = einsum('i,i->', a, a)}
&
\multirow[c]{1}{2.8in}{
    \centering
    \includegraphics[width=1.\linewidth]{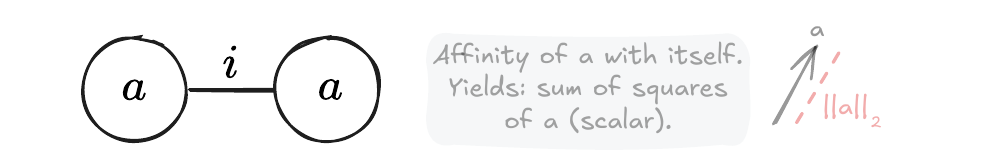}
}
&
{\renewcommand{\arraystretch}{1}$\left[\begin{smallmatrix}\cdot&\cdot&\cdot\end{smallmatrix}\right]\left[\begin{smallmatrix}\cdot\\\cdot\\\cdot\end{smallmatrix}\right] = \cdot$}
&
$h = \sum_i a_i a_i = \|a\|_2$
\\ [.4in]
\hline
 
Normalize \newline \codeitemfoot{asqn = einsum('i,i->', a, a)} \newline
\codeitemfoot{an = a / asqn**0.5}
&
\multirow[c]{1}{2.8in}{
    \centering
    \includegraphics[width=1.\linewidth]{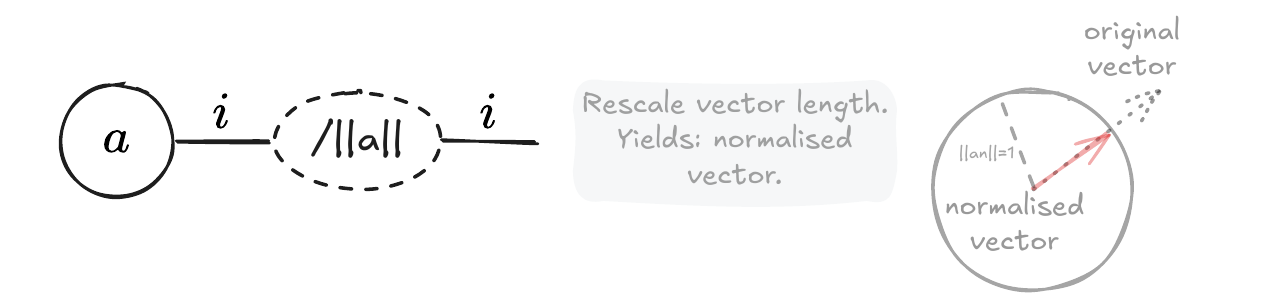}
}
&
{\renewcommand{\arraystretch}{1}$\text{n}\left(\left[\begin{smallmatrix}\cdot\\\cdot\\\cdot\end{smallmatrix}\right]\right) = \left[\begin{smallmatrix}\cdot\\\cdot\\\cdot\end{smallmatrix}\right]$}
&
$\hat a_i = a_i \big/ \sqrt{\sum_j a_j^2}$
\\ [.6in]
\hline

Convex combination \newline \codeitemfoot{h = einsum('i,i->', a, bn)}
&
\multirow[c]{1}{2.8in}{
    \centering
    \includegraphics[width=1.\linewidth]{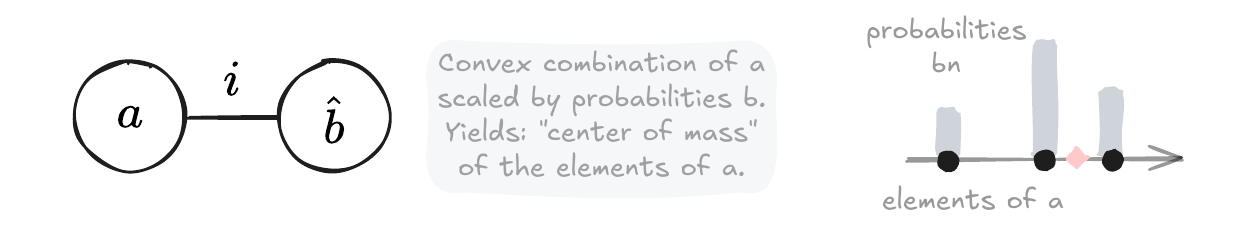}
}
&
{\renewcommand{\arraystretch}{1}$\left[\begin{smallmatrix}\cdot&\cdot&\cdot\end{smallmatrix}\right]\text{n}\left(\left[\begin{smallmatrix}\cdot\\\cdot\\\cdot\end{smallmatrix}\right]\right) = \cdot$}
&
$h = \sum_i a_i \hat b_i$
\\ [.5in]
\hline
 
Cosine similarity \newline \codeitemfoot{h = einsum('i,i->', an, bn)}
&
\multirow[c]{1}{2.8in}{
    \centering
    \includegraphics[width=1.\linewidth]{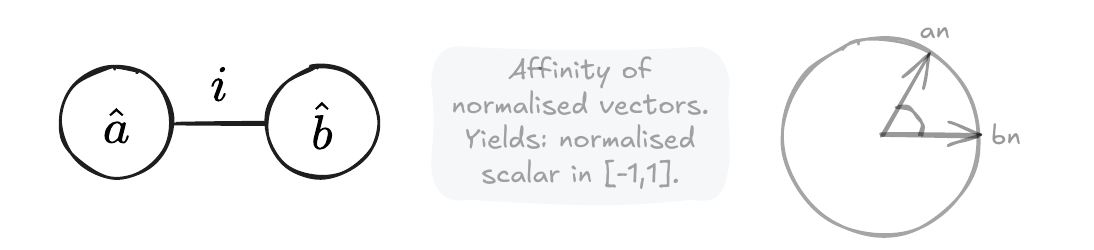}
}
&
{\renewcommand{\arraystretch}{1}$\left[\begin{smallmatrix}\cdot&\cdot&\cdot\end{smallmatrix}\right]\left[\begin{smallmatrix}\cdot\\\cdot\\\cdot\end{smallmatrix}\right] = \cdot$}
&
$h = \sum_i \hat a_i \hat b_i$
\\ [.6in]
\hline

Sum of matrix slices \newline \codeitemfoot{h = einsum('ik->i', A)}
&
\multirow[c]{1}{2.8in}{
    \centering
    \includegraphics[width=1.\linewidth]{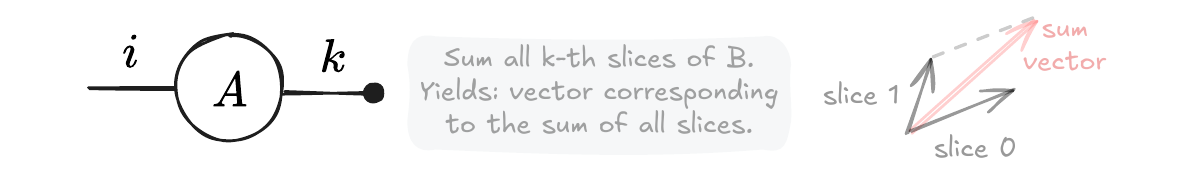}
}
&
{\renewcommand{\arraystretch}{1}$\left[\begin{smallmatrix}\cdot&\cdot\\\cdot&\cdot\\\cdot&\cdot\end{smallmatrix}\right]=\left[\begin{smallmatrix}\cdot\\\cdot\\\cdot\end{smallmatrix}\right]+\left[\begin{smallmatrix}\cdot\\\cdot\\\cdot\end{smallmatrix}\right]$}
&
$h_i = \sum_k A_{ik}$
\\ [.4in]
\hline

Matrix-vector product \newline \codeitemfoot{h = einsum('ij,i->j', A, b)}
&
\multirow[c]{1}{2.8in}{
    \centering
    \includegraphics[width=1.\linewidth]{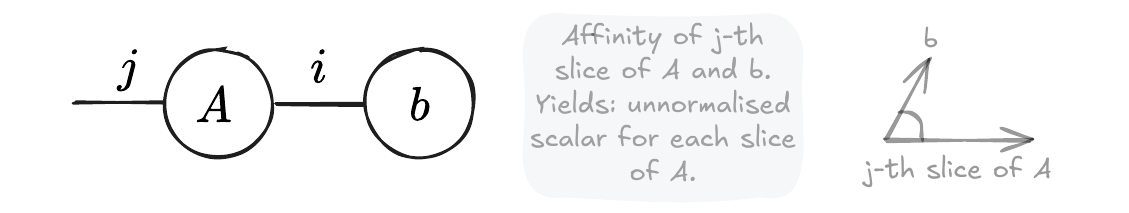}
}
&
{\renewcommand{\arraystretch}{1}$\left[\begin{smallmatrix}\cdot&\cdot&\cdot\\\cdot&\cdot&\cdot\end{smallmatrix}\right]\left[\begin{smallmatrix}\cdot\\\cdot\\\cdot\end{smallmatrix}\right] = \left[\begin{smallmatrix}\cdot\\\cdot\end{smallmatrix}\right]$}
&
$h_j = \sum_i A_{ij}b_i$
\\ [.5in]
\hline

Scaled matrix slices \newline \codeitemfoot{h = einsum('ij,i->ij', A, b)}
&
\multirow[c]{1}{2.8in}{
    \centering
    \includegraphics[width=1.\linewidth]{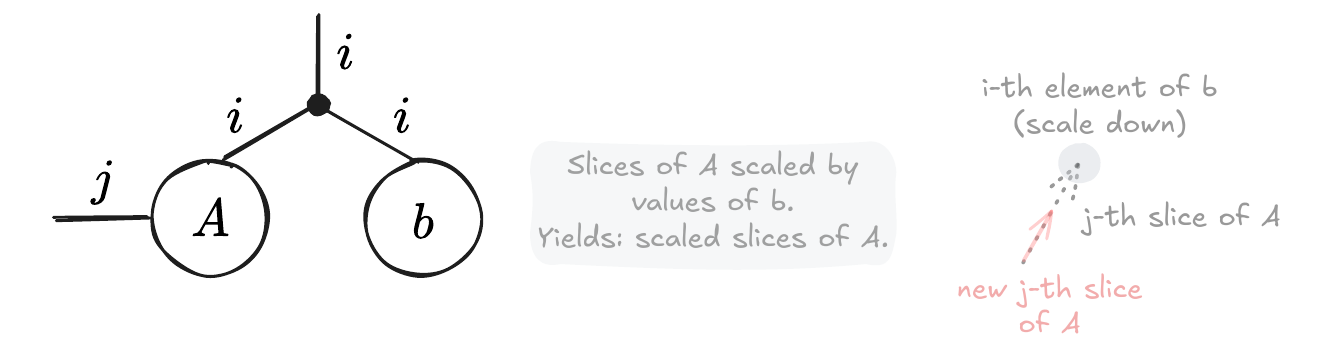}
}
&
--
&
$H_{ij} = A_{ij} b_i$
\\ [.7in]
\hline

Matrix-matrix product \newline \codeitemfoot{H = einsum('ij,ik->jk', A, B)}
&
\multirow[c]{1}{2.8in}{
    \centering
    \includegraphics[width=1.\linewidth]{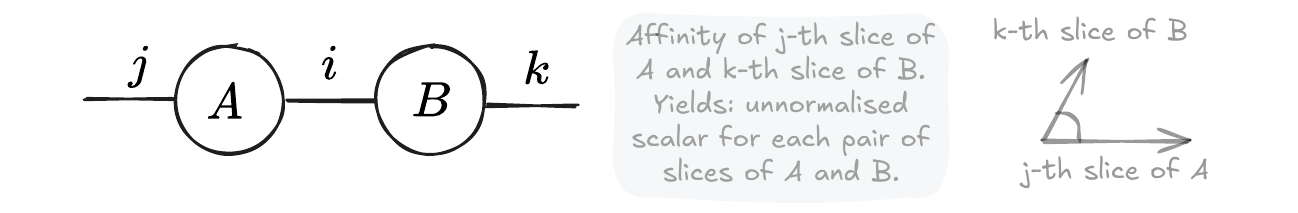}
}
&
{\renewcommand{\arraystretch}{1}$\left[\begin{smallmatrix}\cdot&\cdot&\cdot\\\cdot&\cdot&\cdot\end{smallmatrix}\right]\left[\begin{smallmatrix}\cdot&\cdot\\\cdot&\cdot\\\cdot&\cdot\end{smallmatrix}\right] = \left[\begin{smallmatrix}\cdot&\cdot\\\cdot&\cdot\end{smallmatrix}\right]$}
&
$H_{jk} = \sum_i A_{ij}B_{ik}$
\\ [.4in]
\hline

\end{tabular}
}
\end{table}

\section{Graphical design of simple neural models}
With linear algebra fresh in mind, we can now use the graphical notation to design AI models. As introductory examples, we design two familiar models, a linear model~\citep{berkson1944application,cox1958regression} and a multi-layer perceptron~\citep{rumelhart1986learning}, before moving to more advanced cases, such as self-attention~\citep{vaswani2017attention}.

A linear model~\citep{berkson1944application,cox1958regression} is one of the oldest models in statistics, yet it remains an important baseline for interpretable machine learning, and it forms the backbone of more complex operations in frontier models. A linear model is a matrix-vector product followed by an activation function. The vector $x \in \mathbb{R}^d$ holds the features of an input sample, and the matrix $W \in \mathbb{R}^{h \times d}$ holds the model's learnable parameters. Since this model usually has a non-linear activation which makes the diagram asymmetric, we extend Penrose diagrams drawing the input node in gray and performing tensor operations from left to right (or top-down):
\begin{table}[h]
\centering
\resizebox{\columnwidth}{!}{%
\renewcommand{\arraystretch}{2.4}
\begin{tabular}{
  >{\raggedright\arraybackslash}m{2.1in}
  >{\centering\arraybackslash}m{2.2in}
  >{\centering\arraybackslash}m{1.6in}
}
Linear model \newline 
\codeitemfoot{y = sigma(einsum('j,ij->i', x, W))}
&
\multirow[c]{1}{2.2in}{
    \centering
    \includegraphics[width=1.\linewidth]{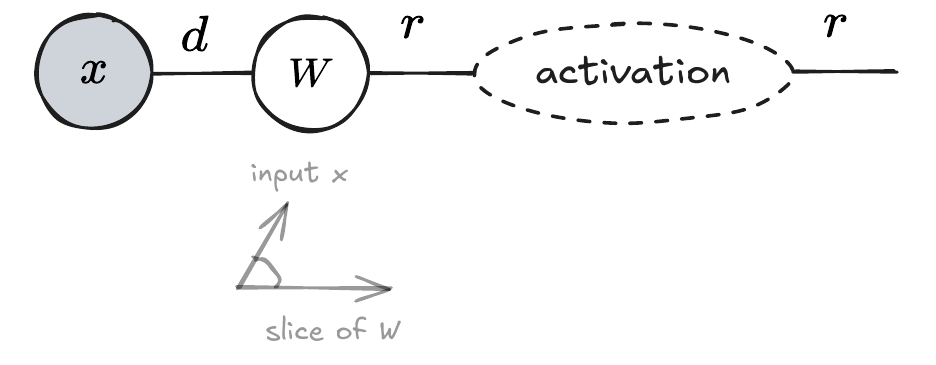}
}
&
$y_i = \sigma \left(\sum_j W_{ij}x_j \right)$ \\ [.4in]
 
\end{tabular}
}
\end{table}

We can apply a linear model to many inputs at once by stacking samples $x_j$ into a batch tensor $X \in \mathbb{R}^{b \times d}$. We can also stack several linear models on top of each other. This gives a multi-layer perceptron (MLP)~\citep{rumelhart1986learning}:
\begin{table}[h]
\centering
\resizebox{\columnwidth}{!}{%
\renewcommand{\arraystretch}{2.4}
\begin{tabular}{
  >{\raggedright\arraybackslash}m{2.5in}
  >{\centering\arraybackslash}m{3.8in}
  >{\centering\arraybackslash}m{2.8in}
}
MLP \newline 
\codeitemfoot{z1 = sigma(einsum('bj,ij->bi', X, W0))} \newline 
\codeitemfoot{z2 = sigma(einsum('bj,ij->bi', z1, W1))} \newline
$\cdots$ \newline
\codeitemfoot{y = sigma(einsum('bj,ij->bi', zL, WL))}
&
\multirow[c]{1}{3.8in}{
    \centering
    \includegraphics[width=1.\linewidth]{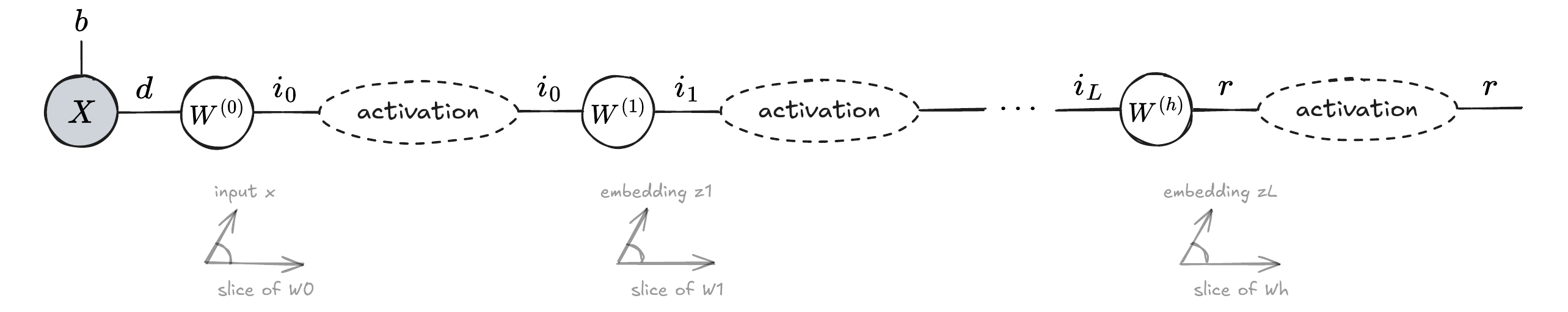}
}
&
$$Y_{bi_L} = \sigma \left( \sum_{i_{L-1}} W_{i_L, i_{L-1}}^{(L)} \dots \sigma \left( \sum_{j_0} W_{i_0, j_0}^{(0)} X_{bj_0} \right) \right)$$
\end{tabular}%
}
\end{table}

Self-attention~\citep{vaswani2017attention} is a key, more complex operation in frontier AI models. This operation projects an input sequence $Z$ of $t$ tokens into a query $q$, key $k$, and value $v$ embeddings. For each pair of tokens $(t,t_p)$, self-attention scores how relevant token $t_p$ is to token $t$. It then uses these relevance scores as weights to combine the value vectors.

Since the tensor manipulations are a bit more complex, we break down the self-attention mechanism into simple atomic manipulations. The first step is to ``copy'' the input $Z$ since we need to reuse this tensor multiple times. In our notation, copying a tensor can be expressed by branching all its legs. We use the index $t_p$ for the legs of the second and third copy of $Z$ as these legs will be used to index key and value tokens the query can attend to:
\begin{figure}[!h]
    \centering
    \includegraphics[width=1\linewidth]{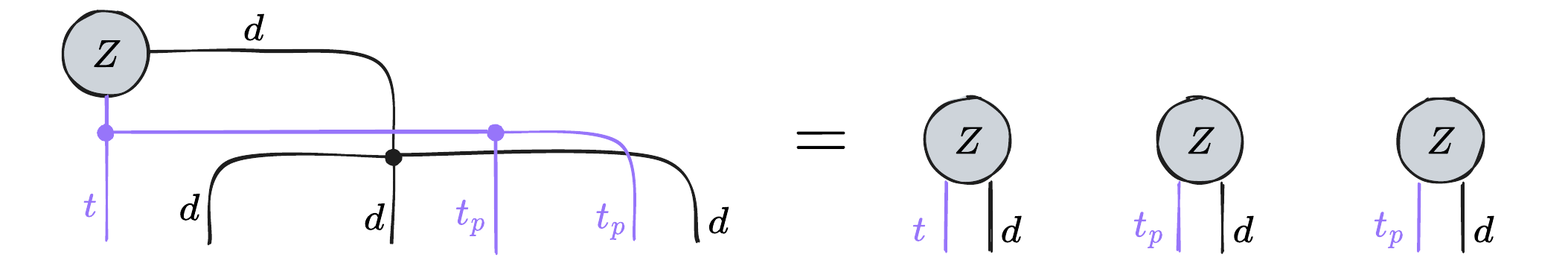}
\end{figure}

Each copy of the tensor $Z$ gets multiplied by a matrix $W \in \mathbb{R}^{d \times e}$ to produce key, query, and value tensors $k,q,v \in \mathbb{R}^{t \times e}$:
\begin{figure}[!h]
    \centering
    \includegraphics[width=0.45\linewidth]{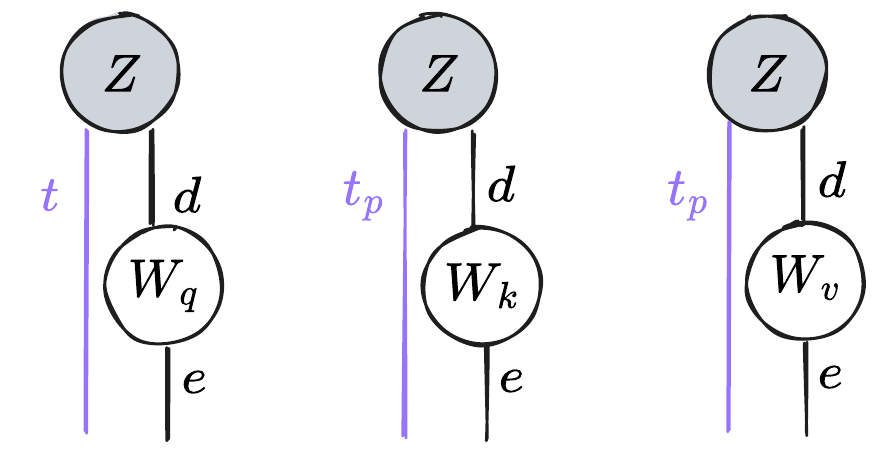}
\end{figure}
\vspace{.3in}

For each pair of tokens $(t,t_p)$, we compute how much the query token $t$ attends to the key token $t_p$:
\begin{figure}[!h]
    \centering
    \includegraphics[width=0.3\linewidth]{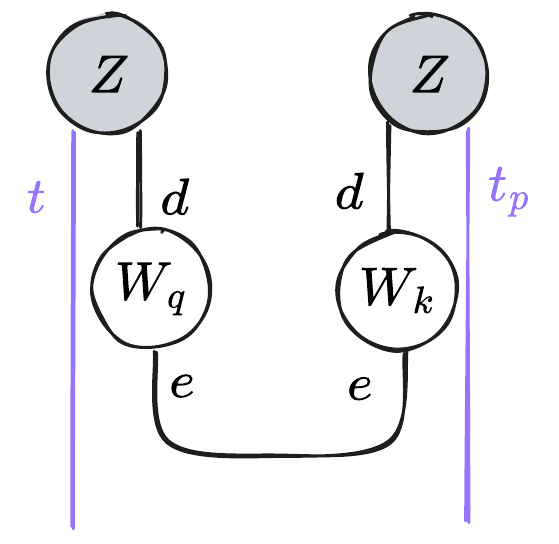}
\end{figure}

We then normalise these ``affinity'' scores into probability values using a softmax activation:
\begin{figure}[!h]
    \centering
    \includegraphics[width=0.3\linewidth]{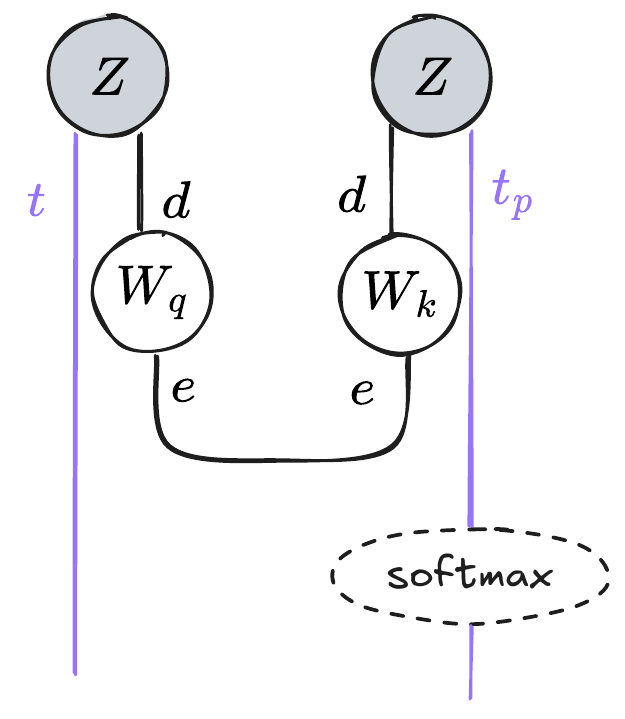}
\end{figure}

And finally we can compute the new embedding of the token $t$ as a convex combination of value embeddings $v$ weighted by their respective probability score:
\begin{figure}[!h]
    \centering
    \includegraphics[width=0.4\linewidth]{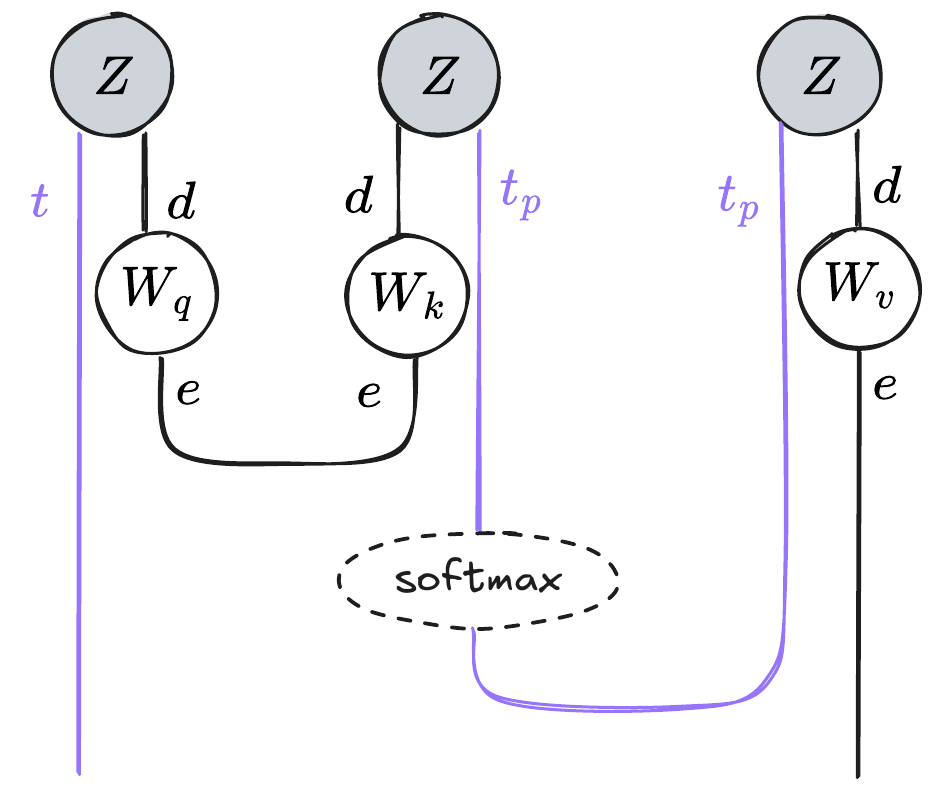}
\end{figure}
\vspace{1.3in}

In a single diagram we can draw self-attention as follows:
\begin{table}[!h]
\centering
\resizebox{\columnwidth}{!}{%
\begin{tabular}{
  >{\raggedright\arraybackslash}m{2.5in}
  >{\centering\arraybackslash}m{4.1in}
  >{\arraybackslash}m{1.4in}
}
Self-attention & 
\multirow[c]{9}{4.1in}{
    \centering
    \includegraphics[width=1.\linewidth]{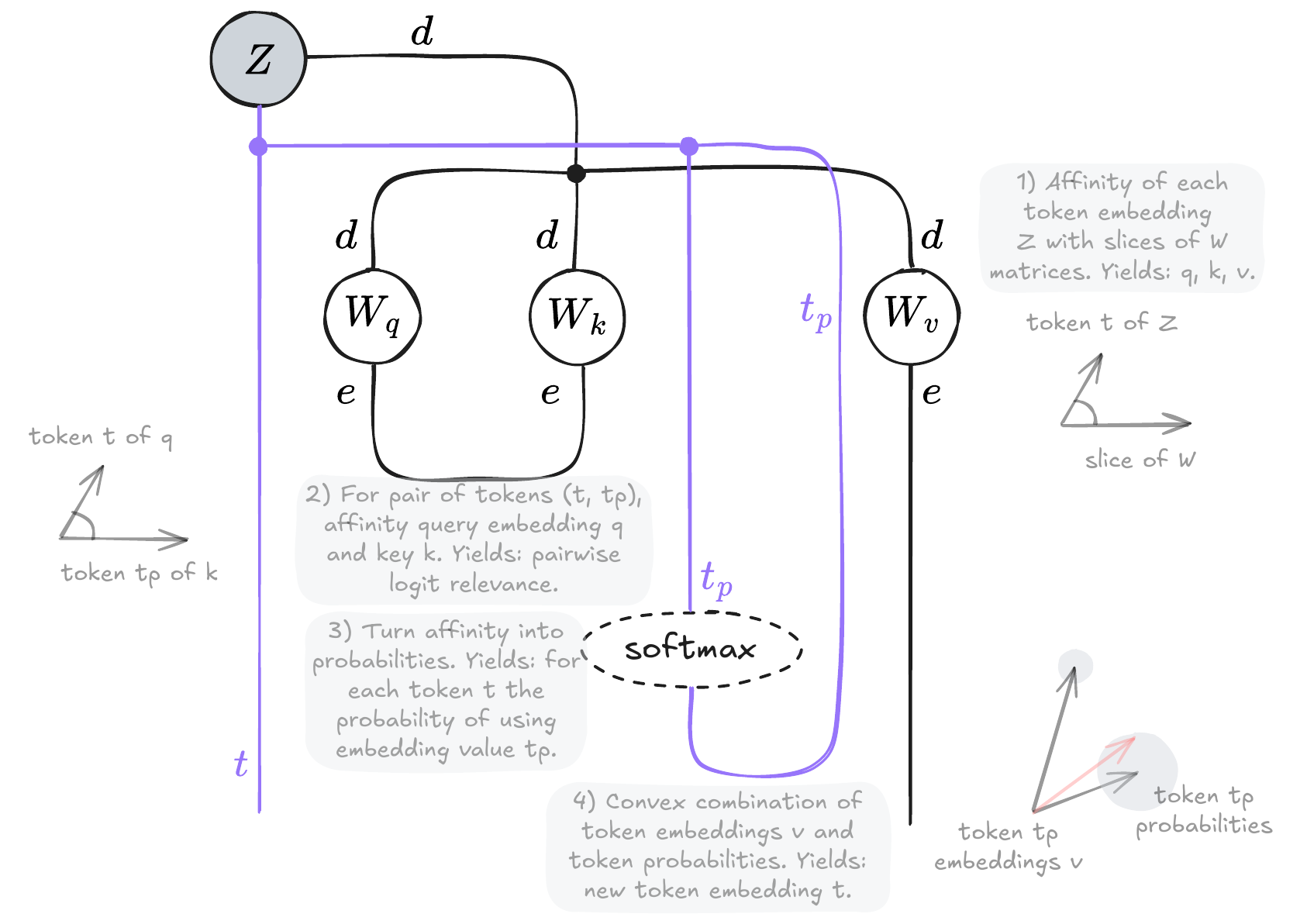}
}
& 
\\ [.7in]

\codeitemfoot{q = einsum('td,de->te', Z, W\_q)} \newline 
\codeitemfoot{k = einsum('pd,de->pe', Z, W\_k)} \newline 
\codeitemfoot{v = einsum('pd,de->pe', Z, W\_v)} 
&
&
$q_{te} = \sum_d Z_{td} W_{q,de}$ \newline 
$k_{pe} = \sum_d Z_{pd} W_{k,de}$ \newline 
$v_{pe} = \sum_d Z_{pd} W_{v,de}$
\\ [.6in]

\codeitemfoot{l = einsum('te,pe->tp', q, k) / sqrt(e)} 
&
&
$l_{tp} = \frac{1}{\sqrt{e}} \sum_e q_{te} k_{pe}$
\\ [.3in]

\codeitemfoot{probs = softmax(l, dim=-1)} 
&
&
$\text{probs}_{tp} = \text{softmax}_p \left( l_{tp} \right)$
\\ [0.4in]

\codeitemfoot{h = einsum('tp,pe->te', probs, v)} 
&
&
$h_{te} = \sum_p \text{probs}_{tp} v_{pe}$
\\ [.3in]

\end{tabular}
}
\end{table}

From here on, diagrams stay minimal: we draw only the indices involved in a contraction. PyTorch supports this directly through ellipsis notation, which lets a tensor operation generalize to any number of batch dimensions. For example, an operation with three preserved indices, batch $b$, token $t$, and head $q$, written as
\begin{center}
    \codeitem{einsum('btqij,btqjk->btqik', A, B)}
\end{center}
can be rewritten as
\begin{center}
    \codeitem{einsum('...ij,...jk->...ik', A, B)}
\end{center}

\section{Graphical design of interpretable architectures}
Interpretable architectures can be generally segmented into three distinct components~\citep{koh2020concept,alvarez2018towards,chen2019looks,barbiero2026standard}: a \emph{backbone} that maps input $x$ to a hidden representation $z$, a \emph{concept encoding map} that turns $z$ into human-meaningful concepts $c$, and a \emph{concept composition map} that turns those concepts into a task prediction $y$. 

Most interpretable architectures use specific tensor operations in their concept encoding and concept composition maps to meet interpretability constraints~\citep{rudin2019stop,barbiero2026standard}. Here we analyse the most common and recurring of these operations, shared across different families of interpretable models.

\subsection{Concept encoding maps}
Concept encoding maps transform latent representations $z$ into representations $c$, known as concepts, that are constrained to align with human semantics. The most common maps in the literature, in order of increasing tensor-manipulation complexity, are probes such as concept activation vectors (CAVs)~\citep{kim2018interpretability} and sparse autoencoders (SAEs)~\citep{ranzato2006efficient,huben2024sparse,templeton2026scaling}, concept bottlenecks~\citep{koh2020concept,espinosa2022concept}, and prototype-based models~\citep{chen2019looks,colamonaco2026prototype}.

\textbf{Sparse encoders}~\citep{ranzato2006efficient} map latent representations $z \in \mathbb{R}^d$ into the sparse activations $c \in \mathbb{R}^k$ via a sparse linear map $W \in \mathbb{R}^{d \times k}$ with $k \gg d$
\begin{table}[h]
\centering
\resizebox{\columnwidth}{!}{%
\begin{tabular}{
  >{\raggedright\arraybackslash}m{2.5in}
  >{\centering\arraybackslash}m{4.1in}
  >{\centering\arraybackslash}m{1.4in}
}
Sparse encoders & 
\multirow{1}{2.5in}{
    \centering
    \includegraphics[width=1.\linewidth]{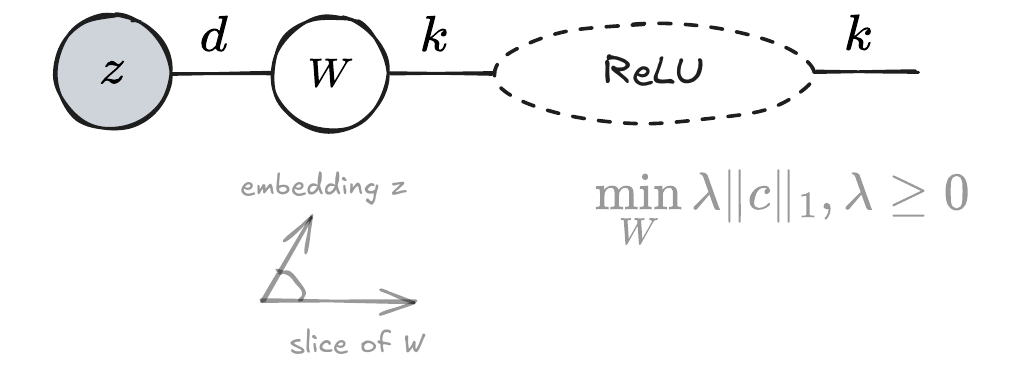}
}
& \\ [.4in]
\codeitemfoot{c = sigma(einsum('d,dk->k', z, W))}
&
&
$c_k = \sigma \left(\sum_d W_{kd}z_d \right)$ \\ [.4in]

\end{tabular}
}
\end{table}

\vspace{1in}

\textbf{Concept bottlenecks}~\citep{koh2020concept} map a latent representation $z$ into the concept representation $c$ via a supervised linear map
\begin{table}[!h]
\centering
\resizebox{\columnwidth}{!}{%
\begin{tabular}{
  >{\raggedright\arraybackslash}m{2.5in}
  >{\centering\arraybackslash}m{4.1in}
  >{\centering\arraybackslash}m{1.4in}
}
Concept bottleneck &
\multirow{1}{2.8in}{
    \centering
    \includegraphics[width=1.\linewidth]{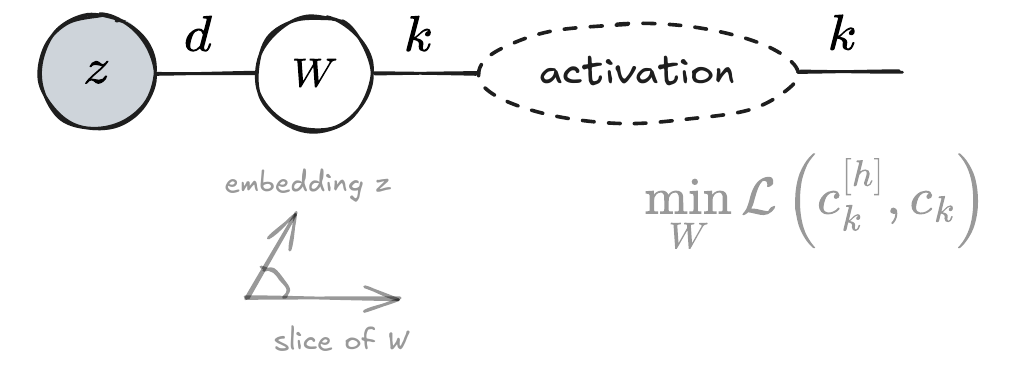}
}
& \\ [.2in] 
\codeitemfoot{c = sigma(einsum('d,dk->k', z, W))}
&
&
$c_k = \sigma \left(\sum_d W_{kd}z_d \right)$ \\[.3in]

\end{tabular}
}
\end{table}

In both cases the tensor operation is identical. The difference lies in the loss and in what the concepts mean: sparse probes recover concept semantics post-hoc (through additional data and labels), while concept bottlenecks build concept semantics into the loss from the start using ground-truth concept annotations $c^{[h]}$.

\textbf{Concept embedding bottlenecks}~\citep{espinosa2022concept} map a latent representation $z$ into a high dimensional concept representation $u \in \mathbb{R}^{d \times k \times s \times e}$ where $s$ is the concept cardinality and $e$ the embedding size. This concept representation is then used to compute concept predictions $c_k$:

\begin{table}[!h]
\centering
\resizebox{\columnwidth}{!}{%
\begin{tabular}{
  >{\raggedright\arraybackslash}m{2.3in}
  >{\centering\arraybackslash}m{4.1in}
  >{\centering\arraybackslash}m{1.3in}
}
Concept embedding bottleneck  & 
\multirow[c]{3}{3.8in}{
    \includegraphics[width=1.\linewidth]{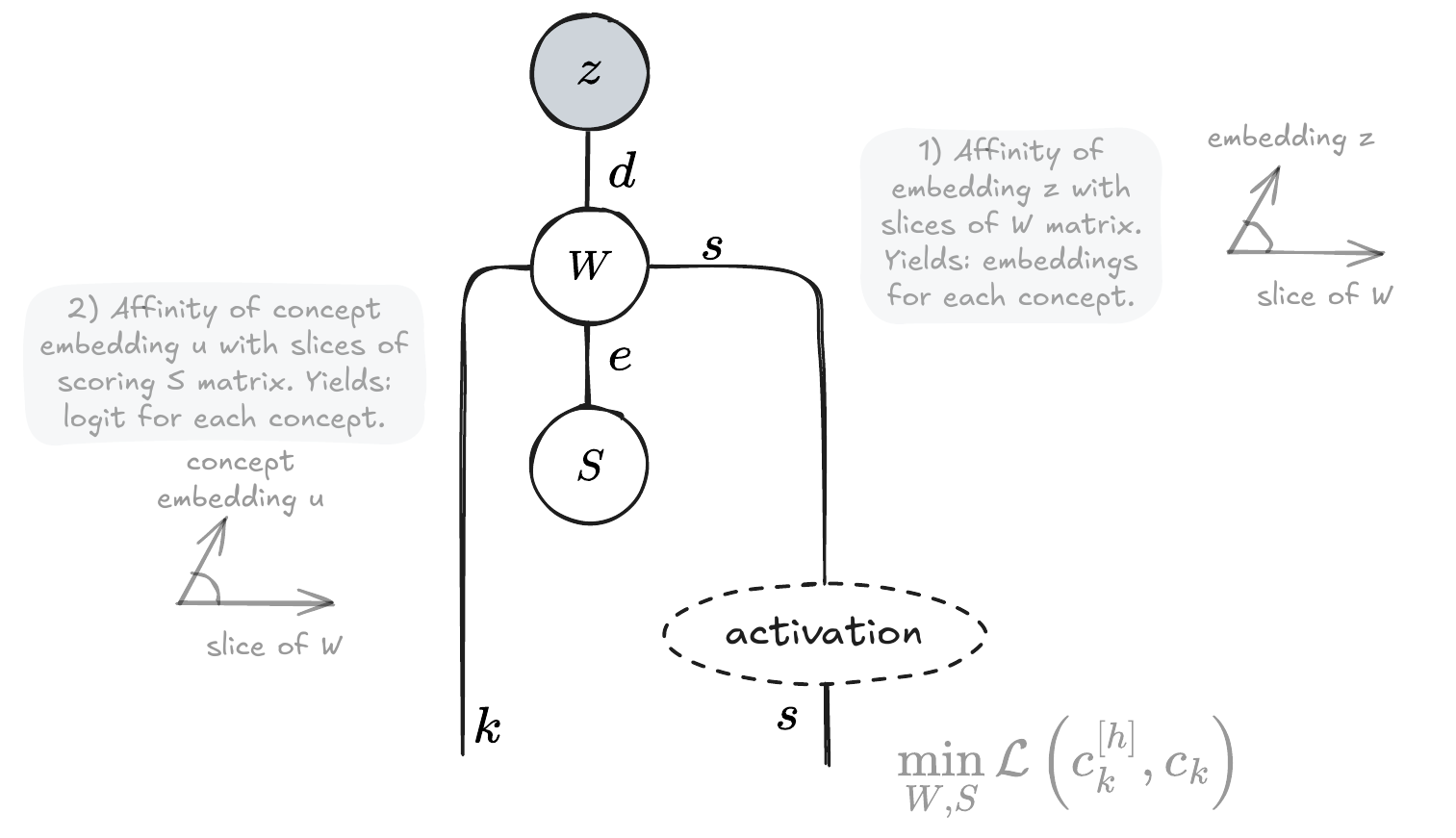}
}
& \\ [.4in] 
\codeitemfoot{u = einsum('d,dkse->kse', z, W)} 
&
&
$u_{kse} = \sum_d W_{dske}z_d$
\\ [.4in]

\codeitemfoot{c = sigma(einsum('kse,e->ks', u, S))}
&
&
$c_{ks} = \sigma \left( \sum_{e} u_{kse} S_{e} \right)$
\\ [.9in]

\end{tabular}
}
\end{table}

\textbf{Prototype-based concept maps}~\citep{chen2019looks,colamonaco2026prototype} need a genuinely different tensor operation. We can think of prototypes as reference examples that tell us whether a concept is active. For instance, the embedding of an apple or a ball can serve as a positive ``prototypical example'' for the concept round, and a fridge or a book as a negative example. Ground-truth prototype labels sit in the tensor $\pi^{[h]} \in \mathbb{R}^{p \times k}$, so each concept $k$ has $p$ labelled prototypes. For a concept $k$ and an input embedding $z \in \mathbb{R}^d$, we compare $z$ against every prototype in $P \in \mathbb{R}^{d \times p \times k}$ and compute the concept label based on input-prototype similarity. For instance, if $z$ is closer to the prototypes for book and fridge than to the prototypes for apple and ball, then the predicted label for round should sit close to $0$. 
\begin{table}[!h]
\centering
\resizebox{\columnwidth}{!}{%
\begin{tabular}{
  >{\raggedright\arraybackslash}m{2.7in}
  >{\centering\arraybackslash}m{4.1in}
  >{\centering\arraybackslash}m{1.2in}
}
Prototype-based concept map & 
\multirow[c]{3}{4.1in}{
    \includegraphics[width=1.\linewidth]{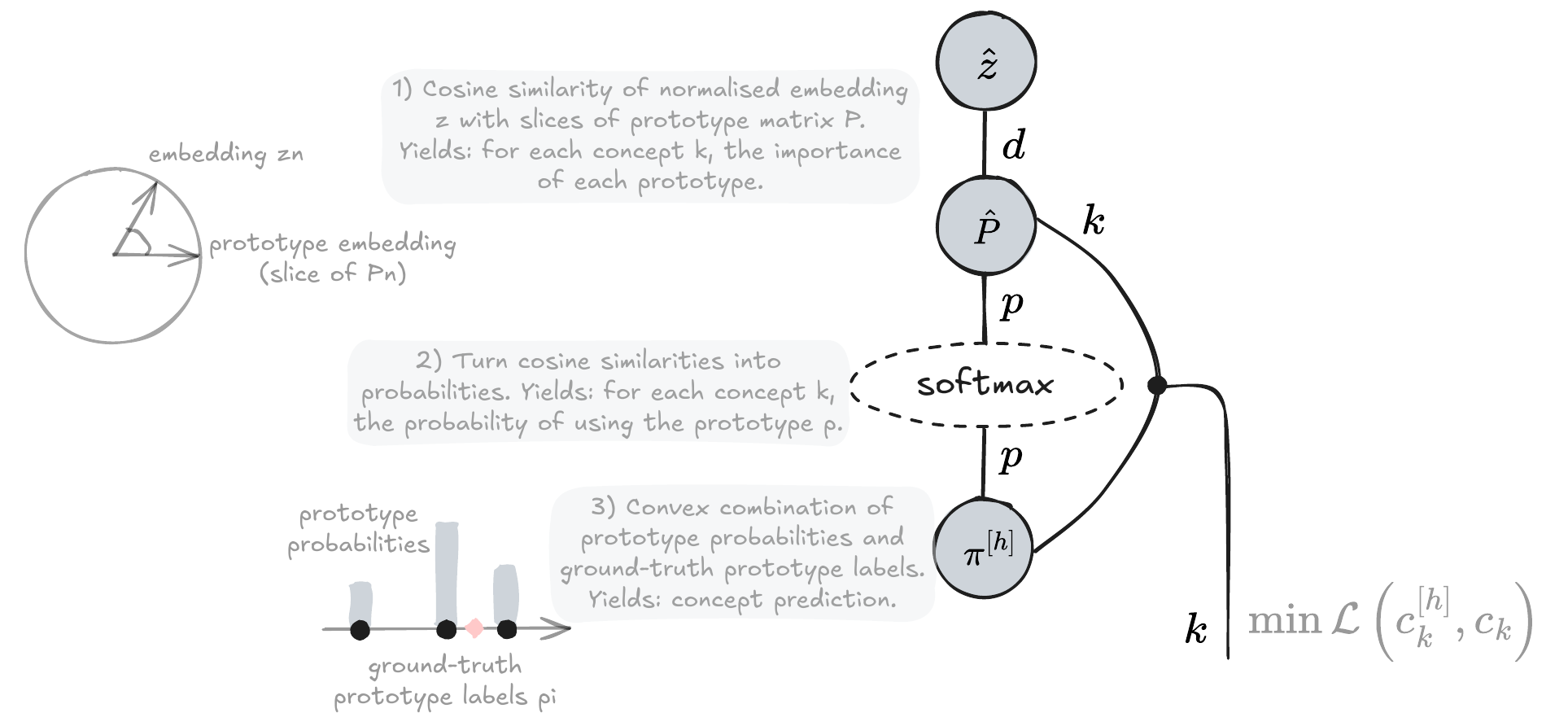}
} 
& \\ [.2in]

\codeitemfoot{l = einsum('d,dpk->pk', zn, Pn)} 
&
&
$\text{l}_{pk} = \sum_d \hat{P}_{dpk}z_d$
\\ [.6in]

\codeitemfoot{probs = softmax(l, dim=-1)} 
&
&
$\text{probs}_{pk} = \text{sm} \left(l_{pk} \right)$
\\ [.4in]

\codeitemfoot{c\_pred = einsum('pk,pk->k', probs, p\_true))}
&
&
$c_k = \sum_p \text{probs}_{pk}\pi_{pk}^{[h]}$
\\

\end{tabular}
}
\end{table}

\subsection{Concept composition maps}
In most interpretability works, the concept composition map is a simple linear model: self-explaining neural nets~\citep{alvarez2018towards}, sparse autoencoders~\citep{huben2024sparse,templeton2026scaling}, concept bottleneck models~\citep{koh2020concept}, all use linear models. A few exceptions are worth discussing: neural additive models~\citep{agarwal2021neural}, concept embedding predictors~\citep{espinosa2022concept}, and mixtures of linear models~\citep{alvarez2018towards,barbiero2023interpretable,debot2024interpretable,desantis2026mixtureconceptbottleneckexperts}.

\vspace{.5in}

\textbf{Neural additive models}~\citep{agarwal2021neural} transform concept activations $c$ independently using a different MLP for each concept $k$ and output task $r$. Then, for each task, they sum the outputs of the MLP of each concept to predict target $y_r$:

\begin{table}[!h]
\centering
\resizebox{\columnwidth}{!}{%
\begin{tabular}{
  >{\raggedright\arraybackslash}m{2.7in}
  >{\centering\arraybackslash}m{4.1in}
  >{\centering\arraybackslash}m{1.7in}
}
Neural additive model & 
\multirow[c]{3}{4.1in}{
    \centering
    \includegraphics[width=1.\linewidth]{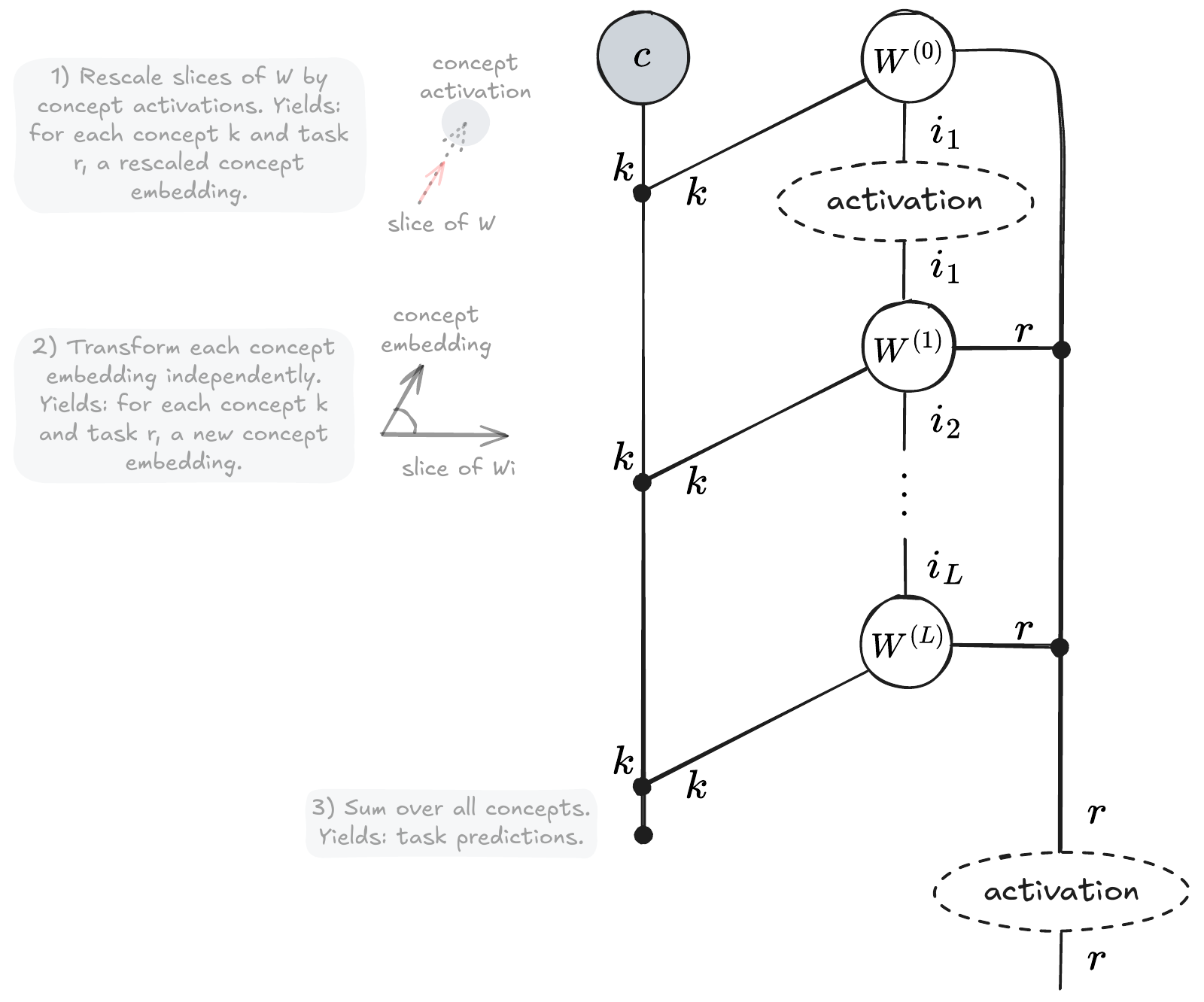}
} 
& \\  [.4in]

\codeitemfoot{h1 = sigma(einsum('k,kri->kri', c, W1))} 
&
&
$h_{kri}^{(1)} = \sigma \left( c_k W^{(1)}_{kri} \right)$
\\ [.6in]

\codeitemfoot{h2 = sigma(einsum('kri,krij->krj', h1, W2))}
&
&
$h_{krj}^{(2)} = \sigma \left( \sum_i h_{kri}^{(1)} W^{(2)}_{krij} \right)$
\\ [.2in]

\dots
&
&
\dots
\\ [.4in]

\codeitemfoot{hL = einsum('kre,kre->kr', he, WL)}
&
&
$h_{kr}^{(L)} = \sum_e h_{kre}^{(L-1)} W^{(L)}_{kre}$
\\ [.5in]

\codeitemfoot{y = sigma(einsum('kr->r', hL))}
&
&
$y_r = \sigma \left(\sum_k h_{kr}^{(L)}\right)$
\\[.4in]

\end{tabular}
}
\end{table}

\textbf{Concept embedding predictors}~\citep{espinosa2022concept} rescale concept embeddings $u$ (e.g., generated by a concept embedding bottleneck) using concept activations $c$ before projecting into the output space $r$:

\begin{table}[!h]
\centering
\resizebox{\columnwidth}{!}{%
\begin{tabular}{
  >{\raggedright\arraybackslash}m{2.in}
  >{\centering\arraybackslash}m{4.1in}
  >{\centering\arraybackslash}m{1.2in}
}
Concept embedding predictor &
\multirow[c]{3}{4.1in}{
    \centering
    \includegraphics[width=1.\linewidth]{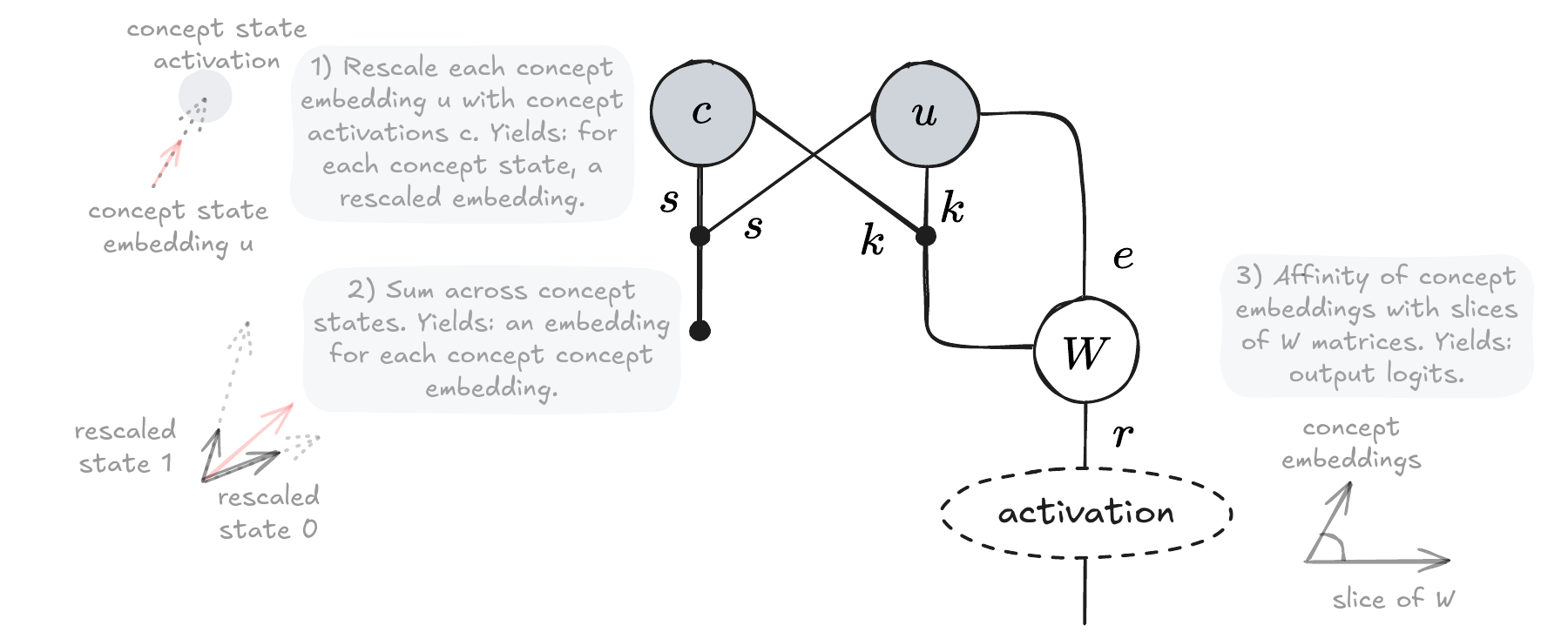}
} 
& \\ [.3in]

\codeitemfoot{h = einsum('kse,ks->kse', u, c)}
&
&
$h_{kse} = u_{kse} c_{ks}$
\\ [.25in]

\codeitemfoot{h = einsum('kse->ke', h)}
&
&
$h_{ke} = \sum_s h_{kse}$
\\ [.1in]

\codeitemfoot{l = einsum('ke,ker->r', h, W)} 
& 
&
$l_r = \sum_{ke} h_{ke} W_{ker}$
\\ [.1in]

\codeitemfoot{l = sigma(l)} 
& 
&
$y_r = \sigma \left(l_r \right)$
\\ [.3in]

\end{tabular}
}
\end{table}

\textbf{Mixtures of linear models}~\citep{desantis2026mixtureconceptbottleneckexperts} compute different predictions for the target $y_r$ using $m$ different linear models. Then each prediction is weighted by the probability of selecting a specific linear model:

\begin{table}[!h]
\centering
\resizebox{\columnwidth}{!}{%
\begin{tabular}{
  >{\raggedright\arraybackslash}m{1.8in}
  >{\centering\arraybackslash}m{4.1in}
  >{\centering\arraybackslash}m{1.2in}
}
Mixture of linear models & 
\multirow[c]{3}{4.1in}{
    \centering
    \includegraphics[width=1.\linewidth]{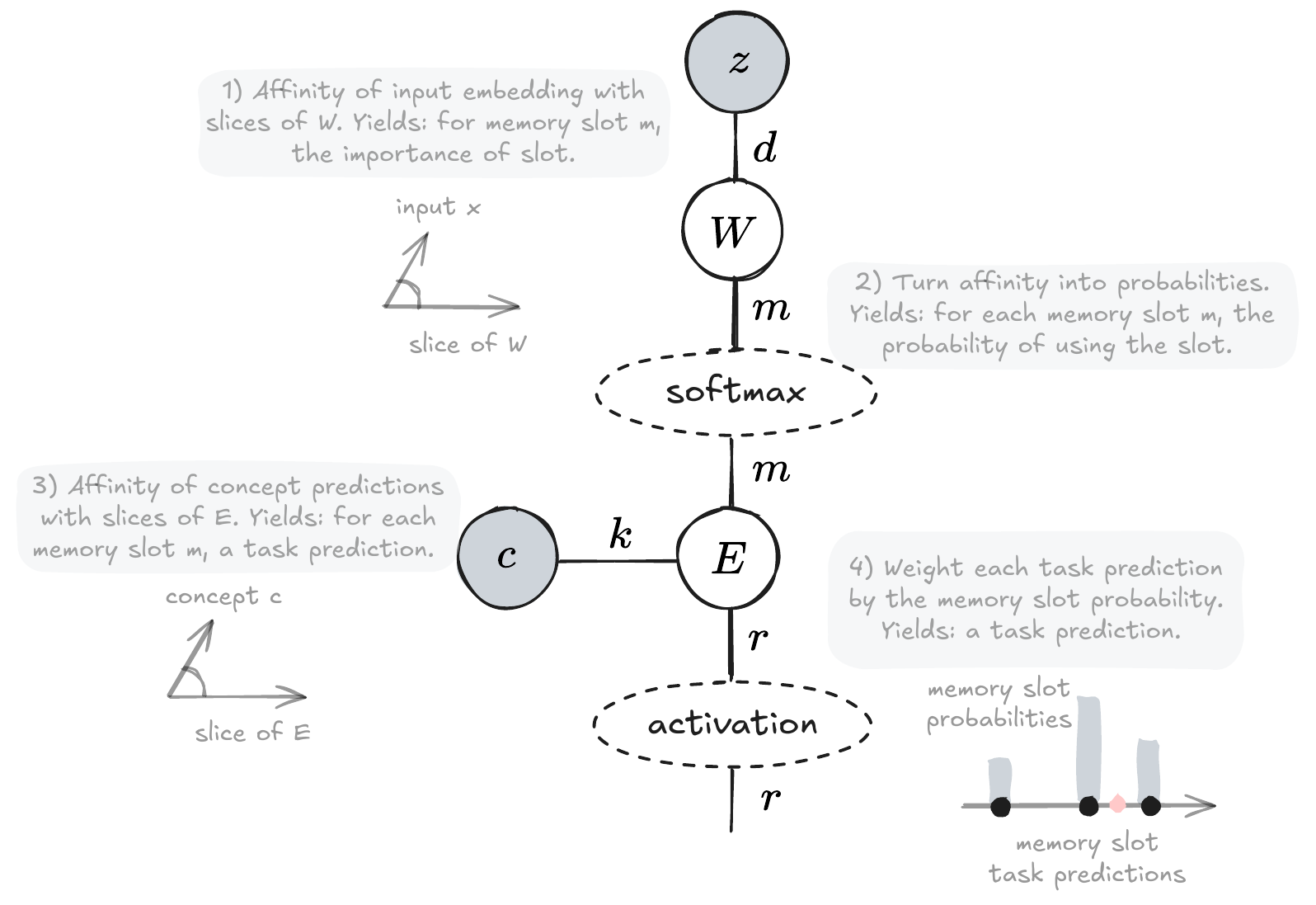}
} 
& \\ [.4in]

\codeitemfoot{l = einsum('d,dm->m', z, W))} 
&
&
$l_m = \sum_d W_{dm} z_d $
\\ [.4in]

\codeitemfoot{pr = softmax(l, dim=-1)} 
&
&
$\text{pr}_m = \text{sm} \left( l_m \right)$
\\ [.4in]

\codeitemfoot{v = einsum('k,kmr->mr', c, E)} 
&
&
$v_{mr} = \sum_k E_{kmr} c_k$
\\ [.2in]

\codeitemfoot{ly = einsum('m,mr->r', pr, v)} 
&
&
$l_r = \sum_{m} v_{mr} \text{pr}_m $
\\ [.2in]

\codeitemfoot{y = sigma(ly)} 
&
&
$y_r = \sigma \left(l_r \right)$
\\ [.4in]

\end{tabular}
}
\end{table}

\vspace{.5in}

\section{Case study: frontier interpretable language models}
As a case study, we diagram the architecture of Steerling-8B~\citep{team2026scaling}, the largest interpretable-by-design language model publicly available at the time of writing.

To keep the focus on the essential tensor manipulations, we drop batch dimensions, since they aren't involved in any contraction, and we show a single attention head; extending to multiple heads is straightforward.
Under these conditions, the essential tensor manipulations in the Steerling-8B architecture take about 30 lines of code.
Drawing the Steerling-8B tensor diagram has three main benefits over the notation used in the original technical report~\citep{team2026scaling}:
\begin{itemize}
    \item It shows at a glance that Steerling-8B is a residual model: the gradient can flow from the output straight back to the first input.
    \item It maps one to one onto PyTorch einops and activation functions, which makes the diagram useful for reproducing the model faithfully.
    \item Every operation in the diagram has a direct geometric reading in linear algebra. This lets us read each manipulation as a transformation in space, which helps build intuition for the underlying computation.
\end{itemize}

\section{Discussion}

\subsection{Related works}
Graphical tensor notation dates to~\citet{penrose1971applications}, who introduced diagrams for tensor contraction in physics. The notation has since been adopted by the categorical-quantum-mechanics community~\citep{coecke2018picturing}, and, more recently, by theoretical computer science and machine learning. 

A first line of work has proposed general-purpose diagrammatic languages for deep learning architectures, without a focus on interpretability.~\citet{chiang2021named} propose named-axis tensor notation to disambiguate operations such as attention.~\citet{abbott2024neural} introduces neural circuit diagrams, a graphical language with a formal correspondence to implementation, later used to derive memory-efficient attention algorithms~\citet{abbott2025flashattentionnapkindiagrammaticapproach}.~\citet{cruttwell2022categorical,lorenz2023causal,gavranovic2024position} pursue a category-theoretic account of architectures more broadly, using string diagrams, a mathematical generalization of Penrose notation, to unify architectures such as convolutional neural nets, recurrent neural nets, and transformers under one algebraic framework.

A more recent line of work started analysing the interpretability literature using graphical notations.~\citet{giannini2024categorical},~\citet{tull2024compositionalinterpretabilityxai}, and~\citet{barbiero2025foundations} use string diagrams to analyse explainable AI methods and interpretable architectures, but without using the tensor manipulation semantics that maps directly to PyTorch programming interfaces.~\citet{taylor2024introduction} applies Penrose notation to mechanistic interpretability, using it to reverse-engineer trained transformer components such as induction heads, building on the informal flowcharts used by~\citet{elhage2021mathematical} to describe transformer circuits. However, this line of work analyses pre-trained opaque models and does not consider tensor manipulations required by inherently interpretable models.

This paper takes a notation developed for post-hoc analysis of trained models and, for the first time to our knowledge, applies this notation to the forward problem: analysing and designing architectures that are interpretable by construction, with a direct, mechanical path from diagram to PyTorch code.

\subsection{Limitations and concrete usage}
Tensor diagrams are exact for multilinear operations, but nonlinearities, masking, and discrete operations such as top-k require ad hoc extensions. Closeness to implementation is also a double-edged sword: diagram size grows with the complexity of the tensor manipulations, so full frontier architectures quickly become unwieldy to draw in conference papers. For this reason, we see tensor diagrams as best paired with a coarser formalism such as probabilistic graphical models. Probabilistic graphical models may be used to capture the high-level causal structure between random variables, while small tensor diagrams specify how each conditional probability function is implemented.

\subsection{Conclusion}
We have shown how tensor diagrams can guide the design of interpretable deep neural networks. For the most common tensor manipulations in interpretability research, we have built a ``Rosetta stone'' showing the matching diagram, PyTorch code, geometric interpretation, and symbolic equation side by side, so readers from different backgrounds can compare and understand them. We also tackled a harder case: we diagrammed and implemented the key modules of a frontier interpretable-by-design language model, Steerling-8B in about 30 lines of code.

Tensor diagrams are expressive, formal, and map directly onto PyTorch code. For these reasons, they could become a standard tool for designing, comparing, and implementing interpretable architectures, alongside other graphical tools such as probabilistic graphical models.

\clearpage
\setpdfpagesize{8.5in}{15.5in}

\begin{table}
\centering
\resizebox{1.\columnwidth}{!}{%
\begin{tabular}{
  >{\raggedright\arraybackslash}m{2.5in}
  >{\centering\arraybackslash}m{4.1in}
}
Steerling-8B & 
\multirow[c]{9}{4.1in}{
    \centering
    \includegraphics[width=1.\linewidth]{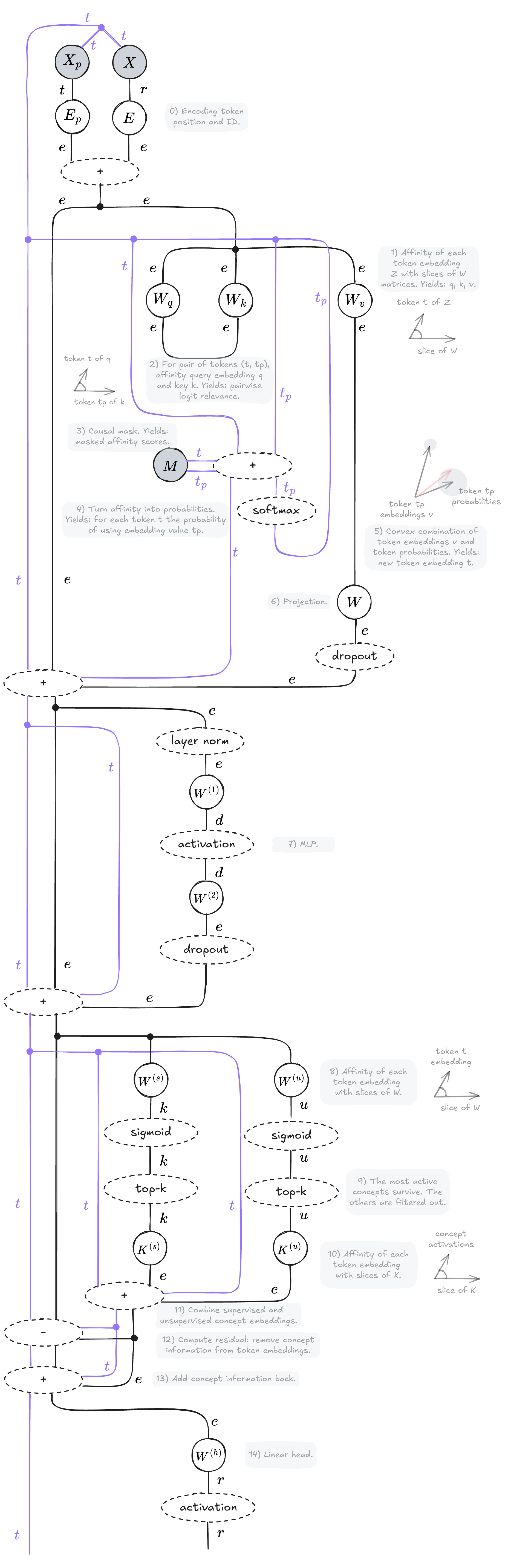}
}
\\ [.6in]

\codeitemfoot{ht = einsum('tr,re->te', X, E)} \newline 
\codeitemfoot{hp = einsum('ti,ie->te', X, Ep)}
&
\\ [.6in]

\codeitemfoot{Z = ht + hp}
&
\\ [.5in]

\codeitemfoot{q = einsum('te,ed->td', Z, W\_q)} \newline 
\codeitemfoot{k = einsum('pe,ed->pd', Z, W\_k)} \newline 
\codeitemfoot{v = einsum('pe,ed->pd', Z, W\_v)} 
&
\\ [.5in]

\codeitemfoot{l = einsum('te,pe->tp', q, k)} \newline
\codeitemfoot{l = l / sqrt(W\_k.shape[1])}
&
\\ [.6in]

\codeitemfoot{l = l + M}
&
\\ [.2in]

\codeitemfoot{probs = softmax(l, dim=-1)} 
&
\\ [0.2in]

\codeitemfoot{h = einsum('tp,pe->te', probs, v)} 
&
\\ [.3in]

\codeitemfoot{h = einsum('te,ed->td', h, W)} \newline
\codeitemfoot{h = dropout(h)}
&
\\ [.4in]

\codeitemfoot{Z = Z + h}
&
\\ [.3in]

\codeitemfoot{h = layernorm(Z)}
&
\\ [.3in]

\codeitemfoot{h = einsum('te,ed->td', h, W1)}
&
\\ [.2in]

\codeitemfoot{h = sigma(h)}
&
\\ [.3in]

\codeitemfoot{h = einsum('te,ed->td', h, W2)}
&
\\  [.2in]

\codeitemfoot{h = dropout(h)}
&
\\ [.4in]

\codeitemfoot{Z = Z + h}
&
\\ [.4in]

\codeitemfoot{ls = einsum('te,ek->tk', Z, Ws)} \newline
\codeitemfoot{lu = einsum('te,eu->tu', Z.detach(), Wu)}
&
\\ [.1in]

\codeitemfoot{cs = sigmoid(ls)} \newline
\codeitemfoot{cu = sigmoid(lu)}
&
\\ [.2in]

\codeitemfoot{csf = topk(cs)} \newline
\codeitemfoot{cuf = topk(cu)}
&
\\ [.2in]

\codeitemfoot{csfe = einsum('tk,ke->te', csf, Ks)} \newline
\codeitemfoot{cufe = einsum('tu,ue->te', cuf, Ku)}
&
\\ [.2in]

\codeitemfoot{ce = csfe + cufe}
&
\\ [.2in]

\codeitemfoot{Z = Z - ce}
&
\\ [.2in]

\codeitemfoot{Z = Z + ce}
&
\\ [.5in]

\codeitemfoot{l = einsum('te,er->tr', Z, Wh)}
&
\\ [.2in]

\codeitemfoot{y = sigma(l)}
&
\\ [.1in]

\end{tabular}
}
\end{table}

\clearpage
\resetpdfpagesize


\bibliographystyle{plainnat}  
\bibliography{references}

\end{document}